\documentclass{article}
\usepackage{graphicx} 
\usepackage{url}
\usepackage{xcolor}

\title{%
  AI in Support of Diversity and Inclusion \\
  
   }

\author{ \c{C}i\c{c}ek  G\"{u}ven
  \and Afra  Alishahi \and
  Henry Brighton \and
 Gonzalo  N\'{a}poles
  \and
  Juan Sebastian Olier
    \and
      Eric Postma
  \and
   Marie \u{S}af\'{a}\u{r}
   \and
   Dimitar Shterionov
  \and
   Mirella  De Sisto
  \and
  Eva Vanmassenhove  
}

\date{\today}

\begin{document}

\maketitle
\begin{abstract}
    In this paper, we elaborate on how AI can support diversity and inclusion and exemplify research projects conducted in that direction. We start by looking at the challenges and progress in making large language models (LLMs) more transparent, inclusive, and aware of social biases. Even though LLMs like ChatGPT have impressive abilities, they struggle to understand different cultural contexts and engage in meaningful, human-like conversations. A key issue is that biases in language processing, especially in machine translation, can reinforce inequality. Tackling these biases requires a multidisciplinary approach to ensure AI promotes diversity, fairness, and inclusion. We also highlight AI’s role in identifying biased content in media, which is important for improving representation. By detecting unequal portrayals of social groups, AI can help challenge stereotypes and create more inclusive technologies. Transparent AI algorithms, which clearly explain their decisions, are essential for building trust and reducing bias in AI systems. We also stress AI systems need diverse and inclusive training data. Projects like the Child Growth Monitor show how using a wide range of data can help address real-world problems like malnutrition and poverty. We present a project that demonstrates how AI can be applied to monitor the role of search engines in spreading disinformation about the LGBTQ+ community. Moreover, we discuss the SignON project as an example of how technology can bridge communication gaps between hearing and deaf people, emphasizing the importance of collaboration and mutual trust in developing inclusive AI. Overall, with this paper, we advocate for AI systems that are not only effective but also socially responsible, promoting fair and inclusive interactions between humans and machines.

\end{abstract}
\section{AI's blessings and threats}

The contemporary advancements in artificial intelligence (AI) have a substantial impact on society. The rise of deep-learning and generative algorithms brings both benefits and risks. On the positive side, they provide new tools that boost productivity and drive scientific breakthroughs. However, these AI algorithms often unintentionally reflect and may even worsen the biases present in the data they are trained on. Moreover, there is a troubling tendency for generative algorithms to be exploited for spreading false information designed to support certain political or commercial interests. Promoting a diverse and inclusive society highlights the need for a careful approach to developing and understanding AI systems. Presently, prominent AI tools such as ChatGPT and Stable Diffusion confront impediments to widespread adoption due to the opaqueness shrouding their training data and fine-tuning methodologies.

The Department of Cognitive Science and Artificial Intelligence (CSAI) at Tilburg University comprises researchers with diverse expertise in computational human cognition and artificial intelligence. This interdisciplinary team provides a unique perspective for addressing the complex challenges related to AI's impact on diversity and inclusion. 

This white paper provides an overview of the research conducted by CSAI researchers and their perspectives on advancing diversity and inclusion within the realm of AI. The selected research covers three broad themes: \emph{transparancy of AI}, \emph{Identifying and resolving biases in AI}, and \emph{Facilitating accessibility and empowering diversity}. In what follows, each of these themes is described in detail.

\section{Transparancy of AI}

Within CSAI researchers try to enhance explainability by uncovering the nature of the internal representations in large language models. 

\subsection*{Delving into the inner workings of Large Language Models - Afra Alishahi} 

The aim of unravelling the opaque nature of large language models (LLMs) is a crucial challenge for their adoption in society. By peering into the proverbial ``black box," researchers try to obtain insights into the mechanisms governing LLM behavior, thereby facilitating the identification and rectification of biases inherent within these systems. 

Recent improvements in the performance of Large Language Models (LLMs) like Galactica and ChatGPT might seem impressive. For a growing number of enterprises and institutions, the efficacy gains made possible by these technologies have been reasons to already deploy them for text and speech processing applications. Yet in various societal contexts, these models still lack the capacity for truly meaningful human-like interactions. To facilitate this, it is necessary to make the social dynamics and characteristics of human communication an integral part of their development.

The responsible deployment of AI and Deep Learning models in general is hindered by the fact that their workings are inherently difficult to interpret and explain. Although academic research has produced an array of techniques for interpreting LLMs, there are as of yet no established methodologies for interpreting and guiding the behavior of these models. This lack of transparency makes it difficult to comprehend the rationale behind LLMs' automated decisions and leaves them open to questions regarding their reliability and trustworthiness. Furthermore, inadequate explanations for users and affected individuals, particularly those of marginalized groups, contribute to the issues of diversity and inclusion.

The contexts in which LLM's are already used often involve diverse user groups, including marginalized communities in care homes, healthcare, and education. It is crucial to acknowledge that LLMs currently lack the capacity for meaningful interaction with humans in such public settings. This limitation is attributed in part to inherent risks, such as the use of stereotypical language or the generation of unfounded and inconsistent statements. The fundamental issue lies in the fact that the current models do not sufficiently consider the social dynamics and characteristics of human communication, and attempts to rectify undesirable behavior are often post-hoc rather than integral to the technology's development.

Fostering successful language interaction between humans and machines requires active engagement in positive and naturalistic communication. Effectively interacting with users in a communicative manner requires awareness and proficiency in understanding diverse socio-cultural backgrounds, social contexts, communicative strategies, and the impact of perceptual context on language interpretation. These considerations surpass the capabilities of Language Models. The field of modern Computational Linguistics should address how to develop computational models that can facilitate human-like interaction and how to implement such models in language technology applications.

\section{Identifying and resolving biases in AI}

Biases are omnipresent in society and therefore also in AI systems. Three lines of CSAI research address these biases. 

\subsection*{Scrutinizing gender biases in language - Eva Vanmassenhove} 

A meticulous examination of gender biases prevalent within linguistic frameworks constitutes another focal point of CSAI research. Through rigorous analysis, researchers aim to unearth latent biases entrenched within language datasets, thereby paving the way for the development of more equitable and inclusive linguistic models

Language technology pervades our daily lives more than ever, via large language models and virtual assistants to automated translations and voice recognition software. If these technologies are biased against certain genders or reinforce harmful stereotypes, they can perpetuate inequality and exclusion.

In this regard, one of the related research areas in the department is how gender biases are currently being propagated via Natural Language Processing technology \cite{vanmassenhove2024gender, lu2023reducing, vanmassenhove2023proceedings, ginelplaying}. This phenomenon is particularly evident in Machine Translation and Multilingual scenarios because different languages express gender in different ways. When Google Translate is used for example to translate a text from Hungarian, a gender-neutral language, into English, it comes up with the following: "She is beautiful. He is clever. [...] He makes a lot of money. She is baking a cake. He's a professor. She is an assistant."

We have been involved in integrating additional features related to gender in order to be able to easily disambiguate different forms, but the initial approach yielded some undesirable side effects. We also have also been working on a rule-based rewriter for gender-neutral forms of English. \cite{vanmassenhove2021neutral} In a separate project in collaboration with Philosophy and Law, we analyse at what point such biases in translation technologies can be considered discriminatory, harmful, or otherwise unethical, and to what extent they may even be unlawful \cite{GrowthEva}.

We value a more multi-disciplinary approach, as opposed to a merely technical one, and believe such an approach can help to turn AI into a force to promote diversity and inclusion. A better, richer understanding of our society (and our own behavior) puts us in a better position to define the fair practices that AI should propagate instead of biases. One can argue that most researchers in this field specialize in computer science, linguistics, sociolinguistics, ethics or law. This is reflected in publications that often present technically advanced solutions, but also show a lack of understanding when it comes to gender across languages and vice versa. AI can then be used to flag biased content, make users aware of potentially biased statements, and empower underrepresented groups by creating tools that address their specific needs.


\subsection*{Mitigating implicit and explicit biases - Gonzalo Napoles}
The development of transparent AI algorithms stands as an imperative in the quest for fostering trust and accountability within AI frameworks. By imbuing AI systems with transparency, researchers aim to engender greater clarity regarding the decision-making processes underlying AI-generated outcomes, thereby mitigating the proliferation of biased outputs. 

While the primary focus of AI research revolves around the accuracy of solutions, there has been a recent shift towards studying the impact of AI-based systems in society. These intelligent systems are as good as the data used to build them, which might contain historical biases reflecting arguably fair and ethical processes and practices. Detecting, mitigating and explaining these explicit or implicit biases is key to building responsible, inclusive AI-based systems that contribute positively to the values of our society. Moreover, the fact that recent scientific breakouts in the AI field have been attached to large private companies like Google, Meta or OpenAI has made fair machine learning research a priority for academicians and governments alike.

One of the main technical challenges of building fair machine learning models is how to reuse existing historical data that might contain biased patterns. Neglecting these data sources is unfeasible since they regularly involve valuable patterns that do not suffer from any bias contamination. To tackle this issue, our team has developed algorithms to detect and mitigate implicit and explicit bias in historical datasets devoted to pattern classification tasks. The first algorithm \cite{napoles2022fuzzy,koutsoviti2021bias} aims to quantify the amount of explicit bias against protected features (such as age, gender, and ethnicity) by using a mathematical theory termed fuzzy-rough sets. The advantage of such an algorithm is that it operates directly with the inconsistent patterns discovered in the data (e.g., a male applicant is more likely to receive a loan than a female applicant even when they have similar factual indicators). Therefore, no additional machine learning algorithm is needed to quantify the effect of problematic patterns on the decisions made by intelligent systems built with the contaminated data. The second algorithm \cite{napoles2021modeling, grau2023measuring} aims to quantify the amount of implicit bias against protected features hidden by unprotected features (such as income, work experience, and education). For example, the unprotected feature ''number of worked years" can be used as a proxy to determine an individual's age, which is a protected feature. The developed algorithm uses a special type of recurrent neural network that explores all possible pathways that connect protected and unprotected features to determine how hidden, biased patterns influence the decisions made from the data. This algorithm is unique in the field and solved a mathematical problem that had remained an open challenge for the scientific community. The third algorithm \cite{patternclassificationbias} aims to mitigate both explicit and implicit bias by solving an optimization problem that discovers what data pieces are responsible for the biased patterns, so they will have less weight when building intelligent systems. Preliminary simulations reported that the algorithm is able to mitigate implicit bias in the data to a large extent (up to 76\% in some cases) with minimum information loss. The advantage of this algorithm compared to other procedures reported in the literature lies in its ability to mitigate explicit and implicit bias simultaneously in a single step.

\section{Facilitating accessibility and empowering diversity}

AI can be utilized to create more inclusive and accessible environments for people of all abilities.       

\subsection*{Diversifying training datasets - \c{C}i\c{c}ek  G\"{u}ven} 
The composition of training datasets wields profound implications for the efficacy and fairness of AI algorithms. In recognition of this salient fact, CSAI researchers advocate for the diversification of training datasets to encompass a broader spectrum of demographic, cultural, and experiential attributes, thereby fostering the development of more inclusive AI frameworks. 

AI's potential to enhance diversity and inclusion becomes evident when it actively identifies and addresses the needs of underprivileged groups, those who are struggling to have access to their basic needs, such as food, healthcare and education. In light of this, we have been focusing on contributing to the solutions of societal problems such as poverty,  malnutrition\cite{mohammedkhan2023image, mohammedkhan2021predicting}, and school dropout \cite{colak2023school}. Collaborating with Tilburg University's Zero Poverty Lab and the NGO Welthungerhilfe (World Hunger Help), we participate in the ''Child Growth Monitor" project. This initiative seeks to identify malnutrition in children through mobile phone-captured images. Successfully achieving this objective would help to enable the identification of at-risk children in remote areas with limited access to healthcare. Our involvement in the project begins with developing AI models that analyze virtual body images to estimate malnutrition-related body parameters, and contributing a dataset of children's images to the field of computer vision \cite{ARAN} to serve the greater goal of Zero Hunger. 

Another ongoing project our researchers are involved in, exemplifying the use of  AI for disadvantaged communities  \cite{iconBrainPoverty}, aims to understand the link between poverty and brain networks. We conduct this under the roof of  Zero Poverty lab,  in an interdisciplinary research environment  consisting of researchers from different schools of Tilburg University. This project  aims to end generational poverty, in collaboration with external stakeholders through nongovernmental organizations such as MOM and Quiet Tilburg. We contribute to the project by working on machine learning techniques in graph representation learning  to investigate the effects of poverty and related variables, on brain structure, function and cognition. Our involvement there will also be around using explainable AI techniques to demystify AI by making the predictions made by AI algorithms more transparent and understandable to everyone, including individuals in poverty.

Effective AI research often involves collaboration across various domains. Interdisciplinary knowledge enables researchers to recognize the complexities and nuances of real-world problems, ensuring that AI solutions are not only technically sound but also practical and beneficial. We benefit from our internally and externally connected and socially aware working culture also for research around diversity and inclusion.
\subsection*{The dynamics of LGBTQ+ disinformation across Europe - Henry Brighton}
In 2021, the European Parliament published a 31-page briefing on disinformation campaigns targeting the European LGBTQ+ community \cite{strand2021disinformation}. The LGBTQ+ community includes individuals who are lesbian, gay, bisexual, transgender, queer, or questioning. An example of disinformation is the claim that the LGBTQ+ community was responsible for the spread of COVID-19. The overall goal of the actors behind anti-LGBTQ+ campaigns is to undermine European unity and destabilize democratic processes in various European countries by attacking “progressive” ideologies that champion diversity and inclusion. A key conclusion of the briefing was that the European Union lacks the infrastructure needed to effectively monitor online disinformation. This is an alarming situation because the technologies underlying the propagation of disinformation, such as social media platforms and search engines, are used by hundreds of millions of people every day. Without effective monitoring, we are blind to what influence these technologies may be exerting, and to what extent disinformation permeates the world wide web.

\begin{figure}[t]
\includegraphics[width=11cm]{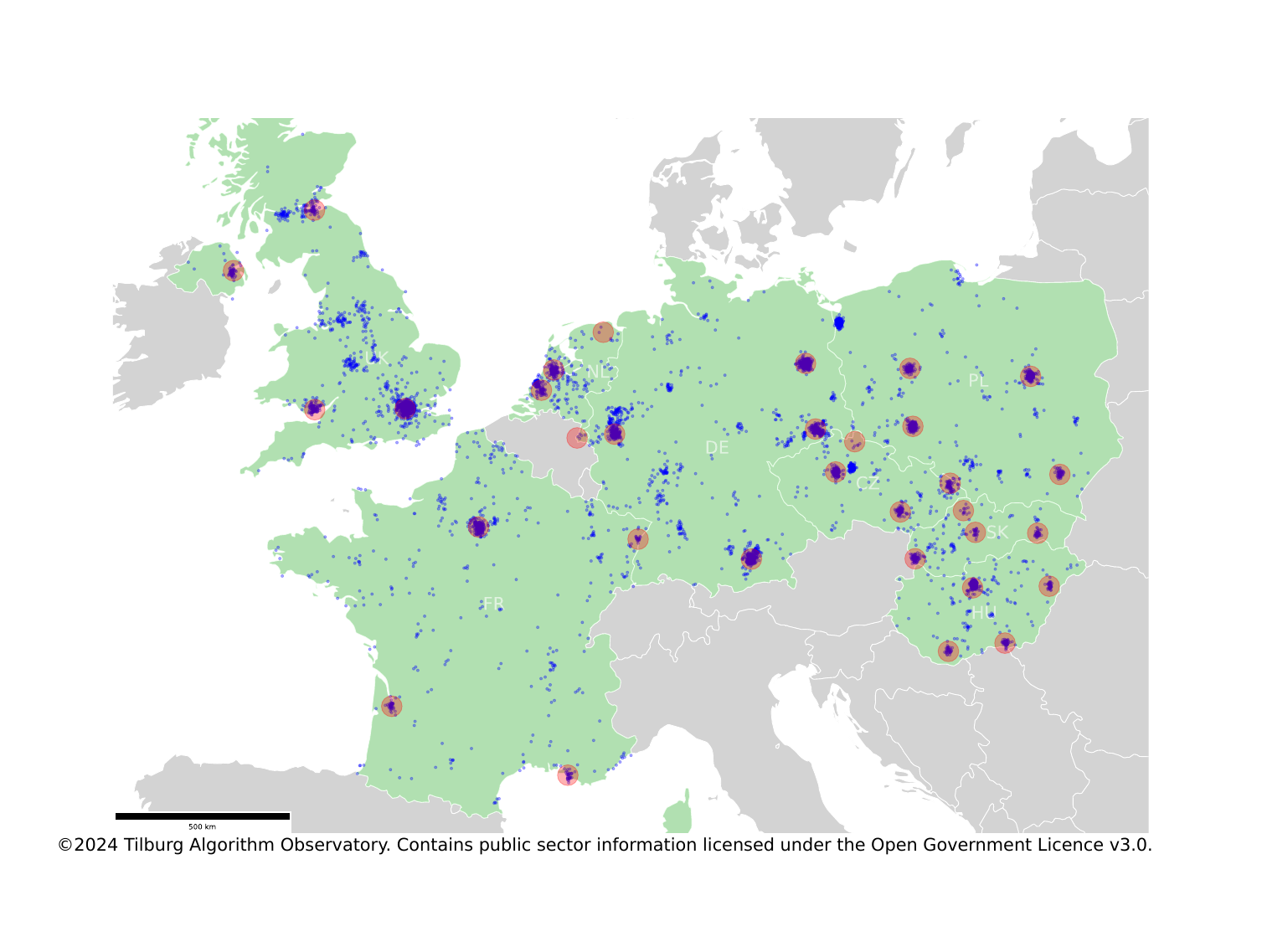}
\centering
\label{fig}
\caption{This map shows the location of residential internet connections we used to monitor LGBTQ+ disinformation in The Netherlands, The United Kingdom, Germany, France, Poland, The Czech Republic, Slovakia, and Hungary. The large red circles show the 24 regional locations where we focused our monitoring, while the small blue circles show the locations of the individual internet connections we used. We have perturbed these locations to protect anonymity.}
\end{figure}

Search Guardian is a project that uses AI techniques to conduct large-scale monitoring of the role of search engines in the propagation of disinformation relating to LGBTQ+ communities in Europe. Led by the Tilburg Algorithm Observatory, this is a collaborative initiative in partnership with the LGBTQ+ community and experts in disinformation and political extremism in Central and Eastern Europe. Monitoring online platforms poses several technical problems, largely because technology companies excel at detecting, and blocking, automated interactions with their services. While short-term and small-scale studies may be easier to implement, the Tilburg Algorithm Observatory focuses on developing resilient monitoring technology that uses artificial intelligence techniques to conduct automated, systematic, and large-scale studies. In our study of disinformation relating to the LGBTQ+ community, we automated over 178k search engine interactions in 8 countries and collected over 1.5 million search results. Figure \ref{fig} shows the location of the residential internet connections used by our monitoring systems for this project.

The scientific and societal impact of this project is best understood by examining how we used our monitoring technology in partnership with the LGBTQ+ community. Indeed, when we refer to the European LGBTQ+ community, we are referring to a number of communities for which variations in language and culture need to be taken into account when studying disinformation narratives. To meet this challenge, we established a collaborative network of partners that included LGBTQ+ experts in The Netherlands, The United Kingdom, Germany, France, Poland, The Czech Republic, Slovakia, and Hungary. Together, we created a multilingual, multi-region dataset of 16k geographically localized search engine queries relating to LGBTQ+ disinformation. Our analysis of 1.5 million search results for these queries revealed that major US search engines prioritize "mass" or "legacy" media outlets, and tend to demote decentralized, peer-to-peer outlets that are more likely to publish extremist views. The non-US independent search engines we monitored, however, are much less likely to mask these alternative media outlets. Our findings raise fundamental questions about the neutrality of web infrastructure and deepen our understanding of disinformation dynamics, particularly in relation to the LGBTQ+ community.

\subsection*{SignON project: Facilitating inclusivity -   Dimitar Shretionov \& Mirella  De Sisto} 

The objective of the 
recently concluded %
SignON project 
was to reduce the communication gap between hearing, deaf and hard-of-hearing people through mobile applications and services for automatic translation between sign and spoken languages  \cite{shterionov2021signon}. But the ``communication gap" is not a one-sided issue and we need to make a clarification here. Reducing the communication gap does not imply “helping deaf people”. It goes in both directions and implies promoting open communication, interactions within the working environment and outside, providing access to information in the preferred language of communication and of course, providing assistive technology that can be used by any of the interlocutors following their mutual agreement. With that respect, we note another issue with technological solutions and “enforced inclusion” (see below).

Within SignON, we 
developed a collection of videos referred to as the SignON encyclopedia that covers a wide range of topics – from deaf culture to neural machine translation.
A fundamental element of our project is Co-creation – a cyclic process that interleaves users’ feedback into development, supporting R\&D and informs the users about the latest advancements. 

\begin{figure}[t]
\includegraphics[width=12cm]{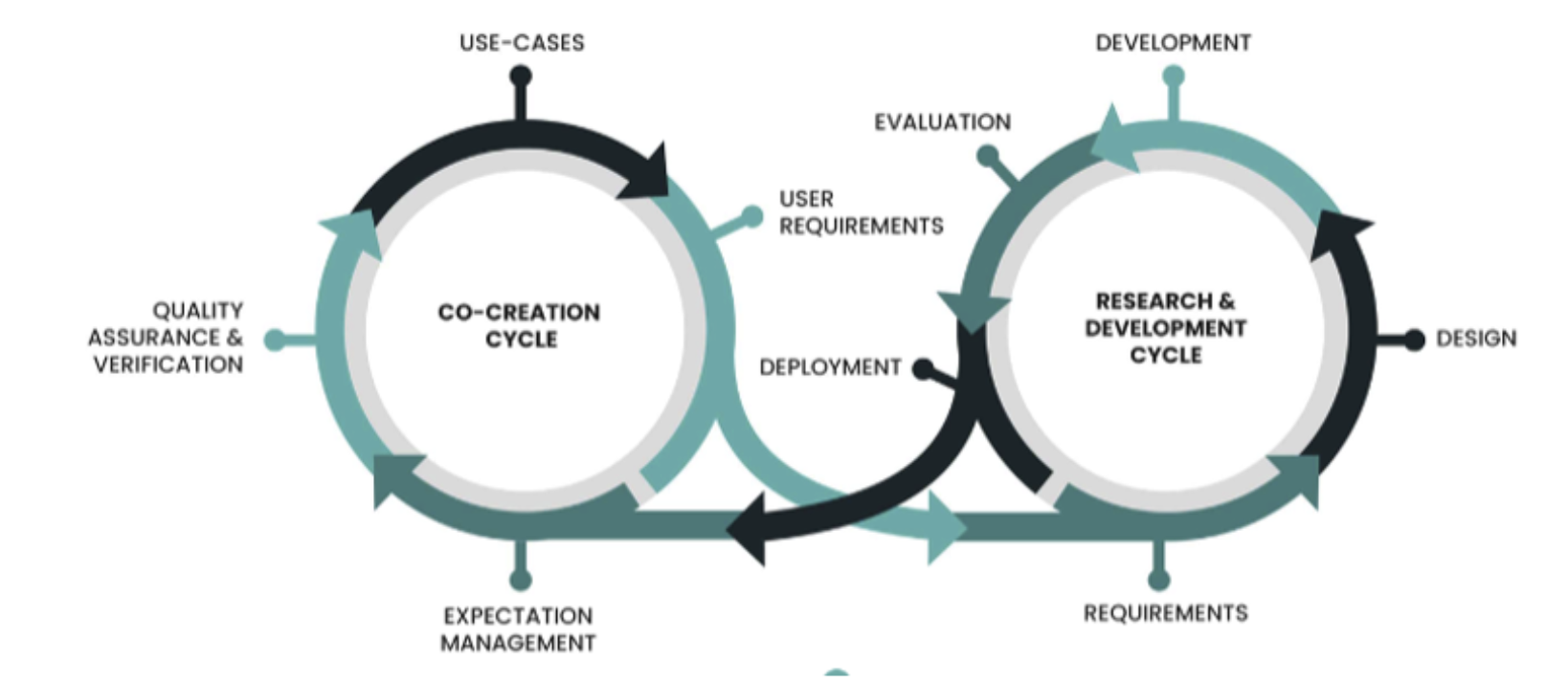}
\centering
\label{fig2}
\caption{Co-creation process}
\end{figure}
As we note above, translation technology has the potential to bring down language (and culture) barriers. However, we need to stress that this does not imply that technology should be the preferred solution for every use case, situation or individual/group. Automatic translation solutions should be part of a repertoire of translation solutions and the decision on which one to use should be in the hands of the interlocutors and not enforced by policymakers. This is even more important when dealing with groups of people that have been highly marginalised in the past. We refer the interested reader to the white paper “Sign Language Technology: Do's and Don'ts, A Guide to Inclusive Collaboration Among Policymakers, Researchers, and End Users.” by Jorn Rijckaert and Davy Van Landuyt (under review).

Through SignON it became clear that sign language data is not enough for the requirements of state-of-the-art neural models. Through additional funding opportunities, we worked on developing two corpora: Gost-Parc-Sign and XSL-HoReCo.

The first one, Gost-Parc-Sign, consists of gold-standard data and aims to give researchers the opportunity to test their models with high-quality data (from “authentic” signers, that is individuals who use a sign language as their main and/or preferred language). The Flemish Sign Language (VGT) videos composing this corpus were produced semi-spontaneously by deaf signers for a deaf audience. Therefore, the quality of the signing is as close as it could possibly be to real life and authentic communication \cite{de2023gost}. Such a testing ground is important for the realistic evaluation of translation solutions between sign and spoken languages.

The second corpus started as a project to collect Sign Language of the Netherlands, written English and Dutch data for the HoReCa domain (i.e. NGT-HoReCo). It was funded under the ELE (European Language Equality 2) project hat; then other SignON members contributed by including more languages to it (VGT, LSE – Spanish Sign Language –, Spanish and Irish) \cite{de2024new}. Such a parallel corpus – with multiple aligned signed and spoken languages – has not yet been published. With the HoReCo corpus, we aim to provide data that researchers can use for the analysis of different languages and the development of (multilingual) systems.

However, the problem of data is not isolated and should be addressed globally. A collaboration between SignON members and EASIER members led to the publication of a white paper aiming at bringing to policymakers' attention the scarcity of language technology resources available for European signed languages \cite{vandeghinste2023report}.

Inclusion, for us, means more than just gathering data and opinions from deaf individuals. It involves deaf, hard of hearing, and hearing people working together as equals, leveraging their diverse backgrounds and expertise to achieve shared goals.

The European Union of the Deaf (EUD) and the Flemish Sign Language Centre (VGTC) 
played significant roles in the SignON consortium. Their involvement 
has promoted collaboration among deaf, hard of hearing, and hearing individuals, fostering knowledge exchange, and building trust within the project.

One research focus in the CSAI department is Inclusive and Sustainable Machine Translation (ISMT). This group explores sustainable and inclusive methods for developing machine translation models. In October 2023, a deaf researcher joined ISMT as a PhD candidate, enriching our research perspective and supporting collaborative efforts. It is important to note that the inclusion of this researcher is not solely for diversity's sake; we equally value their expertise in sign language linguistics, deaf culture, and sign language research. We hope this serves as an example for other researchers and groups to embrace diversity and work together towards common goals.

Between 2021 and 2023, we have offered Dutch Sign Language (NGT) courses in our department to reduce communication barriers and raise awareness. These courses
have been attended by members from both the CSAI department and the Department of Communication and Cognition (DCC). While these courses do not replace the need for interpreters, especially for formal and work-related meetings, they help us communicate better and gain insights into deaf culture. It is important to acknowledge that communicating without interpreters can be cognitively demanding and may result in information loss. Nevertheless, through these initiatives, we strive to find effective ways of communication.

Gaining better insights into how sign languages (SLs) compare to spoken languages can enhance our understanding of cognitive processes. Collaborating with individuals from diverse cultural backgrounds is also an enriching experience.

One motivation for our work is to reduce bias and stigma between different groups. Ultimately, we all need to work together, regardless of our differences. Another motivation is to lead by example. By showcasing successful collaborations between deaf, hard of hearing, and hearing individuals, we hope to inspire others to adopt co-creation processes and build collaborative environments based on trust and shared goals. However, it is important to note that inclusion should not be enforced merely for its own sake.

The increasing demand for data in AI models has led researchers to rely more on synthetic data. This approach can exacerbate biases, particularly affecting minority languages, which may be overshadowed by dominant languages already supported in AI.

\subsection*{Leveraging AI as a tool for bias detection - Juan Sebastian Olier} 
Harnessing the capabilities of AI as a discerning instrument for detecting biased visual and linguistic narratives represents a potent strategy in the pursuit of fostering diversity and inclusion. By deploying AI algorithms endowed with bias-detection capabilities, researchers endeavor to identify and mitigate instances of discriminatory content dissemination. 

The way media represent specific social groups can reinforce prejudices and dominant views about these groups, and can perpetuate their exclusion and lack of power. We seek to illuminate the terminology and frameworks used to describe 'otherness', or how individuals and groups perceive those different from them. By uncovering the issues and logic behind societal categorizations, this research can inform more equitable and just systems for recognizing 'the other' that do not sustain exclusion.

AI offers a powerful means to study and detect these issues on a large scale. Similarly, AI can uncover new societal issues, raising awareness of exclusionary issues. In a previous project for example we used Deep Learning techniques for an automatic visual content analysis of how migrants are portrayed by media in ten countries. We compared the general group 'migrants' with the specific groups 'refugees' and 'expats'. We found that portrayals predominantly focus on asylum seekers and associate them with poverty and risks for host societies. The results also showed that the demographics as portrayed in the images deviate from official statistics to various degrees per country. In the portrayal of expats for instance 'white' faces are overrepresented, while 'Asian' faces are underrepresented. And confirming previous research, we found that migrant women are underrepresented and portrayed as younger than men \cite{olier2022stereotypes}.

Currently, we are working on understanding the predominant views of the other from a nationalist perspective. The 'other' is often defined by their nationality, and this definition shapes perspectives and gives rise to stereotypes and discriminatory behaviors. Recognizing and challenging these stereotypes is crucial to understanding broader phenomena associated with continuing specific dynamics of, for example, power asymmetries that can affect individuals solely due to their nationality. In the future, the goal will focus on the automatic discovery of biases in the representation of social groups in media images. The objective is to better understand societal stereotypes and prejudices while developing novel bias discovery methods that can be applied to large image sets.

Such insights are crucial for developing more responsible and trustworthy AI approaches, auditing existing AI systems, or assessing their potential societal impacts. More responsible AI requires a better understanding of the phenomena associated with defining the other. Disregarding those phenomena in developing AI systems - which happens because identity-related attributes are often considered insignificant, indisputable and apolitical - these systems can amplify societal issues and perpetuate exclusion. On the other hand, a clear understanding of the issues in question contributes to AI systems that can challenge the normative frameworks that sustain biased views of 'others' \cite{scheuerman2020we}. As such, AI can be a force for tackling societal issues regarding inequality and exclusionary practices.

\section{The broader picture}

CSAI research encompasses more than the examples described in this white paper. In particular, its research addresses other societal relevant themes, such as AI for biodiversity, AI for zero hunger, AI for zero poverty, AI for energy and water infrastructure, AI for cybersecurity, and AI for disaster management and defense. 

In conclusion, this paper has explored how AI can support diversity and inclusion by highlighting the importance of transparent, socially responsible, and context-aware approaches in AI research. While large language models show promise, they also have limitations—especially in understanding cultural nuances and reducing social biases—which calls for a multidisciplinary approach. Projects like Child Growth Monitor and SignON demonstrate how AI can address real-world issues like malnutrition and accessibility. This paper also explores how AI models can be applied in collaboration with both the affected community and domain experts, as demonstrated by the Tilburg Algorithm Observatory’s work on disinformation targeting the LGBTQ+ community. Achieving these goals will require inclusive datasets, trust-building, and ensuring that AI algorithms make fair, clear decisions. As AI develops, fairness and inclusivity must remain priorities to prevent reinforcing stereotypes and inequalities. Through these efforts, we aim to build AI systems that not only work effectively but also promote a fairer, more inclusive society.

\bibliography{mybib}{}

\begin{thebibliography}{10}

\bibitem{colak2023school}
Hazal Colak~Oz, {\c{C}}i{\c{c}}ek G{\"u}ven, and Gonzalo N{\'a}poles.
\newblock School dropout prediction and feature importance exploration in malawi using household panel data: machine learning approach.
\newblock {\em Journal of Computational Social Science}, 6(1):245--287, 2023.

\bibitem{de2024new}
Mirella De~Sisto, Vincent Vandeghinste, Caro Brosens, Myriam Vermeerbergen, and Dimitar Shterionov.
\newblock {XSL-HoReCo} and {GoSt-ParC-Sign: Two New Signed Language - Written Language Parallel Corpora}.
\newblock In {\em Selected papers from the {CLARIN Annual Conference} 2023}, pages 23--33, 2024.

\bibitem{de2023gost}
Mirella De~Sisto, Vincent Vandeghinste, Lien Soetemans, Caro Brosens, and Dimitar Shterionov.
\newblock {GoSt-ParC-Sign}: {Gold Standard Parallel Corpus of Sign} and spoken language.
\newblock In {\em Proceedings of the Annual Conference of the European Association for Machine Translation}, pages 503--504, 2023.

\bibitem{ginelplaying}
Mar{\'\i}a Isabel~Rivas Ginel, Sarah Theroine, {\'E}ric Poirier, Lo{\"\i}c Barrault, Felix Hoberg, Oliver Czulo, Eva Vanmassenhove, Izabella Thomas, and Lucie Bernard.
\newblock Playing with gender. all-inclusive games machine translation: The all-ingmt project.
\newblock In {\em Book of Abstracts}, page 144, 2022.

\bibitem{grau2023measuring}
Isel Grau, Gonzalo N{\'a}poles, Fabian Hoitsma, Lisa~Koutsoviti Koumeri, and Koen Vanhoof.
\newblock Measuring implicit bias using shap feature importance and fuzzy cognitive maps.
\newblock In {\em Intelligent Systems Conference}, pages 745--764. Springer, 2023.

\bibitem{patternclassificationbias}
Fabian Hoitsma, Gonzalo N{\'a}poles, {\c{C}}i{\c{c}}ek G{\"u}ven, and Yamisleydi Salgueiro.
\newblock Mitigating implicit and explicit bias in structured data without sacrificing accuracy in pattern classification.
\newblock {\em AI \& SOCIETY}, pages 1--20, 2024.

\bibitem{koutsoviti2021bias}
Lisa Koutsoviti~Koumeri and Gonzalo N{\'a}poles.
\newblock Bias quantification for protected features in pattern classification problems.
\newblock In {\em Progress in Pattern Recognition, Image Analysis, Computer Vision, and Applications: 25th Iberoamerican Congress, CIARP 2021, Porto, Portugal, May 10--13, 2021, Revised Selected Papers 25}, pages 351--360. Springer, 2021.

\bibitem{lu2023reducing}
Tianshuai Lu, No{\"e}mi Aepli, Annette Rios, Eva Vanmassenhove, Beatrice Savoldi, Luisa Bentivogli, Joke Daems, and Jani{\c{c}}a Hackenbuchner.
\newblock Reducing gender bias in nmt with fudge.
\newblock In {\em In: 1st Workshop on Gender-Inclusive Translation Technologies (GITT)}. University of Zurich, 2023.

\bibitem{mohammedkhan2021predicting}
Hezha MohammedKhan, Marleen Balvert, {\c{C}}i{\c{c}}ek G{\"u}ven, and Eric Postma.
\newblock Predicting human body dimensions from single images: a first step in automatic malnutrition detection.
\newblock In {\em CAIP 2021: Proceedings of the 1st International Conference on AI for People: Towards Sustainable AI, CAIP 2021, 20-24 November 2021, Bologna, Italy}, page~48. European Alliance for Innovation, 2021.

\bibitem{mohammedkhan2023image}
Hezha MohammedKhan, {\c{C}}i{\c{c}}ek G{\"u}ven, Marleen Balvert, and Eric Postma.
\newblock Image-based body shape estimation to detect malnutrition.
\newblock In {\em Proceedings of SAI Intelligent Systems Conference}, pages 577--590. Springer, 2023.

\bibitem{ARAN}
Hezha MohammedKhan, Cascha van Wanrooij, Eric Postma, {\c{C}}i{\c{c}}ek G{\"u}ven, Marleen Balvert, Chenar Omer Ali AL~JAF, and Heersh~Raof Saeed.
\newblock Aran: Age-restricted anonymized dataset of children images and body measurements, 2024.
\newblock submitted.

\bibitem{napoles2021modeling}
Gonzalo N{\'a}poles, Isel Grau, Leonardo Concepci{\'o}n, Lisa~Koutsoviti Koumeri, and Jo{\~a}o~Paulo Papa.
\newblock Modeling implicit bias with fuzzy cognitive maps.
\newblock {\em arXiv preprint arXiv:2112.12713}, 2021.

\bibitem{napoles2022fuzzy}
Gonzalo N{\'a}poles and Lisa~Koutsoviti Koumeri.
\newblock A fuzzy-rough uncertainty measure to discover bias encoded explicitly or implicitly in features of structured pattern classification datasets.
\newblock {\em Pattern Recognition Letters}, 154:29--36, 2022.

\bibitem{olier2022stereotypes}
Juan~Sebastian Olier and Camilla Spadavecchia.
\newblock Stereotypes, disproportions, and power asymmetries in the visual portrayal of migrants in ten countries: an interdisciplinary ai-based approach.
\newblock {\em Humanities and Social Sciences Communications}, 9(1):1--16, 2022.

\bibitem{scheuerman2020we}
Morgan~Klaus Scheuerman, Kandrea Wade, Caitlin Lustig, and Jed~R Brubaker.
\newblock How we've taught algorithms to see identity: Constructing race and gender in image databases for facial analysis.
\newblock {\em Proceedings of the ACM on Human-computer Interaction}, 4(CSCW1):1--35, 2020.

\bibitem{shterionov2021signon}
Dimitar Shterionov, Vincent Vandeghinste, Horacio Saggion, Josep Blat, Mathieu De~Coster, Joni Dambre, Henk Van~den Heuvel, Irene Murtagh, Lorraine Leeson, and Ineke Schuurman.
\newblock The signon project: a sign language translation framework.
\newblock In {\em 31st Meeting of Computational Linguistics in The Netherlands (CLIN 31)}, 2021.

\bibitem{strand2021disinformation}
Cecilia Strand, Jakob Svensson, Roland Blomeyer, and Margarita Sanz.
\newblock {\em Disinformation campaigns about LGBTI+ people in the EU and foreign influence}.
\newblock European Parliament, Policy Department for External Relations, 2021.

\bibitem{iconBrainPoverty}
{Tilburg University}.
\newblock From impoverished to enriched brains: A lifespan perspective on the link between poverty, nutrition, cognitive functioning and underlying brain networks.
\newblock \url{https://www.tilburguniversity.edu/about/digital-sciences-society/projects/impoverished-enriched-brains}, 2024.
\newblock online, accessed at 24 04 2024.

\bibitem{GrowthEva}
{Tilburg University}.
\newblock An interdisciplinary analysis of gender-based discrimination in translation technology.
\newblock \url{ https://www.tilburguniversity.edu/about/digital-sciences-society/projects/gender-based-discrimination-translation-technology}, 2024.
\newblock online, accessed at 24 04 2024.

\bibitem{vandeghinste2023report}
Vincent Vandeghinste, Mirella De~Sisto, Maria Kopf, Marc Schulder, Caro Brosens, and Lien Soetemans.
\newblock Report on europe's sign languages. ele project deliverable 1.40.
\newblock {\em Report on Europe’s Sign Languages}, 2023.

\bibitem{vanmassenhove2024gender}
Eva Vanmassenhove.
\newblock Gender bias in machine translation and the era of large language models.
\newblock {\em arXiv preprint arXiv:2401.10016}, 2024.

\bibitem{vanmassenhove2021neutral}
Eva Vanmassenhove, Chris Emmery, and Dimitar Shterionov.
\newblock Neutral rewriter: A rule-based and neural approach to automatic rewriting into gender-neutral alternatives.
\newblock {\em arXiv preprint arXiv:2109.06105}, 2021.

\bibitem{vanmassenhove2023proceedings}
Eva Vanmassenhove, Beatrice Savoldi, Luisa Bentivogli, Joke Daems, and Jani{\c{c}}a Hackenbuchner.
\newblock Proceedings of the 1st workshop on gender-inclusive translation technologies.
\newblock In {\em First Workshop on Gender-Inclusive Translation Technologies at the European Association for Machine Translation Conference}, 2023.

\end{thebibliography}
\bibliographystyle{plain}

\end{document}